\documentclass{article}
\usepackage[OT1]{fontenc}
\usepackage{iclr2027_conference,times}
\usepackage{amsmath,amssymb}
\usepackage{booktabs}
\usepackage{colortbl}
\usepackage{placeins}
\usepackage{float}
\floatstyle{ruled}
\newfloat{algorithm}{tbp}{loa}
\floatname{algorithm}{Algorithm}
\usepackage[hidelinks]{hyperref}
\usepackage{url}
\usepackage{tikz}
\usetikzlibrary{arrows.meta,positioning,fit,calc}
\definecolor{resultblue}{RGB}{228,241,248}

\newif\ifshowrevisions
\showrevisionsfalse
\newcommand{\revision}[1]{\ifshowrevisions\textcolor{red}{#1}\else#1\fi}

\newif\ifshowtodayrevisions
\showtodayrevisionsfalse

\title{Seg2Click: Learning Pixel-Level Action Masks for GUI Visual Grounding}

\author{Jingchao Wang \qquad Dingjiang Huang\thanks{Corresponding author.} \\
\normalfont School of Data Science and Engineering \\
\normalfont East China Normal University, Shanghai 200062, China \\
\texttt{jcwang@stu.ecnu.edu.cn} \quad \texttt{djhuang@dase.ecnu.edu.cn}}
\hypersetup{
  pdftitle={Seg2Click: Learning Pixel-Level Action Masks for GUI Visual Grounding},
  pdfauthor={Jingchao Wang and Dingjiang Huang}
}
\iclrfinalcopy

\begin{document}
\maketitle
\fancyhead{}

\begin{abstract}
Visual grounding is a central challenge for GUI agents, requiring an instruction to be mapped to a target within a crowded screenshot. Coordinate generation with single-point supervision reduces a spatially extended target to one position, despite the presence of many valid click locations. Region-based patch scoring alleviates this ambiguity, but its spatial decisions remain tied to visual-token footprints that may span the target, neighboring controls, and background. To address these limitations, we propose Seg2Click, which reformulates GUI grounding as executable-region prediction: identifying where a click can validly execute the given instruction. Using bounding-box supervision, Seg2Click predicts an executable-region mask and computes the final click point through a deterministic rule. This formulation captures the spatial extent of valid click locations and enables finer localization beyond the visual-token grid, making it well suited to small, densely arranged controls without requiring a verifier or zoom-in pass. Experiments across five backbone configurations from two model families demonstrate the effectiveness of Seg2Click. Seg2Click-9B achieves 92.06\%, 94.10\%, and 55.22\% on ScreenSpot, ScreenSpot-v2, and ScreenSpot-Pro, respectively, while matched-backbone comparisons show improvements over both coordinate generation and patch-based grounding.
Code will be available at \url{https://github.com/jcwang0602/Seg2Click}.
\end{abstract}

\section{Introduction}
\label{sec:introduction}

GUI agents carry out user instructions across web, mobile, and desktop environments~\citep{zhou2024webarena,rawles2025androidworld,xie2024osworld}. A key capability is visual grounding: locating the target specified by an instruction within a screenshot~\citep{cheng2024seeclick,gou2025uground,qin2025uitars}. This task is particularly challenging in dense professional interfaces, where small controls are often placed close together~\citep{li2025screenspotpro}. Recognizing the relevant toolbar or control group is insufficient: successful interaction requires placing the click within the specific element named by the instruction.

Existing approaches commonly address this localization problem through either direct coordinate generation or relevance prediction over visual tokens. Coordinate-generation methods represent screen positions as text and directly predict a click location~\citep{chen2023shikra,cheng2024seeclick,wu2025osatlas}. Under single-point supervision, these methods are trained to reproduce one annotated coordinate, even though many locations within the target may be equally valid. For example, different clicks inside a button can trigger the same action, whereas a nearby click just outside its boundary may activate another control. Single-point supervision does not explicitly capture this distinction, because it penalizes deviations from the annotated point regardless of whether they remain within the target. This mismatch motivates predicting the valid target region before selecting a specific click location. Visual-token-based methods move toward region prediction by assigning relevance scores to image patches. In particular, GUI-Actor supervises all visual patches that overlap the target, allowing multiple patches to receive positive supervision~\citep{wu2025guiactor}. Although this approach relaxes the commitment to a single annotated point, its spatial decisions remain tied to the visual-token grid. A single token footprint may cover part of the target, neighboring controls, and background, yet receive only one relevance score. Interpolating these scores increases output resolution but does not introduce independently learned predictions within each footprint. Thus, moving from point supervision to region supervision leaves a granularity bottleneck unresolved, especially for small, densely arranged controls. The remaining challenge is to preserve region-based supervision while learning spatial decisions finer than the visual tokens that represent the screenshot.

To address this challenge, we propose Seg2Click, which reformulates GUI grounding as executable-region prediction: predicting the region in which a click can validly execute the instruction, then computing the click point using a deterministic rule. We instantiate this formulation as box-supervised sub-token action-field prediction, learning distinct scores within each visual-token footprint (Figure~\ref{fig:seg2click_overview}). A query-conditioned pixel-shuffle decoder~\citep{shi2016espcn,wang2025mllmseg} combines aligned vision and language-model features to produce an executable-region mask at the processed screenshot resolution. Target bounding boxes provide supervision, and the centroid of the mask's largest connected component determines the final click. This formulation addresses the two limitations above: region supervision accommodates multiple valid click locations, while sub-token prediction enables finer spatial distinctions within token footprints. It requires no additional contour annotations, verifier, or zoom-in pass.

\begin{figure}[t]
\centering
\includegraphics[width=\linewidth]{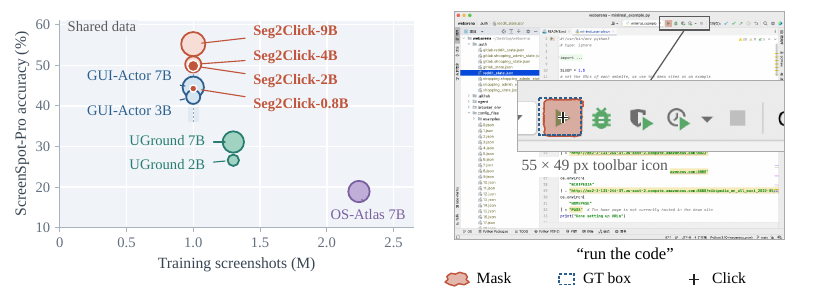}
\caption{\textbf{Left:} ScreenSpot-Pro accuracy versus training screenshots; marker area denotes backbone size. Qwen3.5-based Seg2Click and GUI-Actor share the same grounding data, aligned at ${\sim}1$M screenshots. \textbf{Right:} Qwen3.5-4B predicts the mask and click for ``run the code.'' The display-only inset shows the $55\times49$ pixel icon and its ground-truth box.}
\label{fig:seg2click_teaser}
\end{figure}

\begin{figure}[t]
\centering
\includegraphics[width=\linewidth]{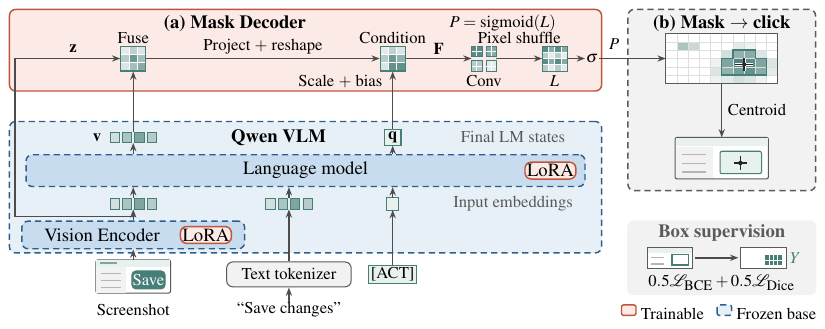}
\caption{Overview of Seg2Click. (a) Aligned vision and language-model features are fused and conditioned on an action query to predict distinct scores within each visual-token footprint, supervised by masks derived from target boxes. (b) A fixed rule converts the predicted mask into a click at the centroid of its largest connected component. Orange denotes trainable modules; blue dashed outlines indicate frozen backbone weights.}
\label{fig:seg2click_overview}
\end{figure}

To assess whether these changes translate into more accurate grounding, we first compare Seg2Click with coordinate generation and patch scoring, then evaluate its applicability across backbones. Under matched backbone, training data, and optimization settings, Seg2Click-0.8B produces 24 more correct clicks than coordinate generation on ScreenSpot-Pro. On the same 0.8B backbone, it also outperforms reproduced GUI-Actor by 4.87 percentage points, corresponding to 77 additional correct clicks (Table~\ref{tab:output_representation}). Beyond these comparisons, experiments cover five backbone configurations from two model families. Seg2Click achieves 47.00\% on ScreenSpot-Pro with Qwen2.5-VL-3B and 55.22\% with Qwen3.5-9B, supporting the effectiveness of executable-region prediction across model families and scales (Figure~\ref{fig:seg2click_teaser}).

Our contributions are threefold:
\begin{itemize}
    \item We reformulate GUI grounding as executable-region prediction, separating the learning of valid click regions from deterministic click selection.
    \item We introduce Seg2Click, which learns box-supervised sub-token action fields to resolve spatial distinctions within each visual-token region, without additional contour annotations, a verifier, or a zoom-in pass.
    \item We validate Seg2Click across five backbone configurations from two model families. It outperforms coordinate generation under matched training and reproduced patch scoring on the same backbone, with the largest gain over patch scoring on ScreenSpot-Pro. Our 9B model achieves the highest accuracy among the compared systems on all three benchmarks, while feature and decoder analyses support the design choices.
\end{itemize}

\section{Related Work}
\label{sec:related_work}

\paragraph{GUI understanding and coordinate-based grounding.}

GUI agents map user instructions to interface elements through visual understanding and grounding. UIBert, Ferret-UI, and OmniParser advance UI representation and understanding~\citep{bai2021uibert,you2024ferretui,lu2024omniparser}, while Mind2Web, WebVoyager, Aguvis, and ShowUI support learning and evaluating visually grounded interaction~\citep{deng2023mind2web,he2024webvoyager,xu2024aguvis,lin2024showui}. A common grounding formulation directly generates target coordinates from screenshots and instructions. SeeClick, OS-Atlas, Aria-UI, and TinyClick improve this formulation through specialized models and training strategies~\citep{cheng2024seeclick,wu2025osatlas,yang2025ariaui,pawlowski2025tinyclick}, while UGround and Jedi broaden grounding coverage through scalable data construction~\citep{gou2025uground,xie2025jedi}. Seg2Click addresses a complementary limitation: single-point supervision does not explicitly capture multiple valid click locations within a target. It instead predicts a box-defined action region before deriving the final coordinate.

\paragraph{Attention- and region-based grounding.}

Spatial relevance prediction offers an alternative to direct coordinate generation. TAG aggregates pretrained multimodal attention to localize GUI targets without fine-tuning~\citep{xu2025tag}. GUI-Actor learns an attention-based action head with multi-patch supervision and an optional verifier for candidate selection~\citep{wu2025guiactor}. Seg2Click builds on region-based grounding by refining prediction granularity: a single patch score cannot distinguish target and non-target locations within the same token footprint. Its spatial decoder instead learns distinct action scores within each footprint, and a fixed largest-component centroid rule converts the predicted mask into a click, without a verifier or zoom-in pass.

\paragraph{Language-conditioned dense prediction.}

Language-conditioned segmentation connects multimodal understanding with dense prediction. LISA decodes segmentation-token representations into masks, while PixelLM uses a segmentation codebook and pixel decoder for multi-target reasoning~\citep{lai2024lisa,ren2024pixellm}. MLLMSeg combines visual and language-model features with pixel-shuffle decoding~\citep{wang2025mllmseg}. Seg2Click adapts these techniques to GUI grounding through dynamic-resolution grids and an action-placeholder query, predicting a box-defined action mask for deterministic click extraction. Unlike BoxInst, which uses boxes as weak supervision to recover object masks~\citep{tian2021boxinst}, Seg2Click directly learns the region defined by the benchmark's box-hit criterion, without requiring contour annotations.

\section{Seg2Click}
\label{sec:method}

\subsection{Action-Mask Formulation}
GUI grounding maps a screenshot $I$ and an instruction $u$ to a click location. Under the box-hit evaluation protocol, multiple locations within the annotated target box $b$ are equally valid. Seg2Click represents this spatial target as a dense \textbf{action mask}, separating region prediction from click selection. The annotated box defines the supervision region and serves as a proxy for the executable area in static grounding.

Let $I'\in\mathbb{R}^{H'\times W'\times3}$ be the screenshot after the backbone's dynamic-resolution preprocessing. We predict
$$
P_\theta(I',u)=\sigma\!\left(L_\theta(I',u)\right)\in[0,1]^{H'\times W'},
\qquad \hat{\mathbf c}=D(P_\theta(I',u)),
$$
where $L_\theta$ is a dense logit field, $\sigma$ is sigmoid, and $D$ is a deterministic rule that returns a normalized click. Pixel-level prediction refers to the processed screenshot resolution. As shown in Figure 2, Seg2Click combines aligned visual features with an instruction-aware action query, predicts distinct scores within each visual-token footprint, and converts the resulting mask into a single click.

\subsection{Query-Conditioned Visual Features}
The backbone represents the processed screenshot on an $h\times w$ visual-token grid. At grid location $i$, we extract the post-merge vision feature $\mathbf z_i$ and the final language-model state $\mathbf v_i$ at the corresponding image-token position. Both lie in $\mathbb{R}^d$ and share the same spatial correspondence. We combine these two representations through residual fusion:
\begin{equation}
\mathbf f_i=\mathbf v_i+\operatorname{MLP}\!\left([\mathbf v_i;\phi(\mathbf z_i)]\right),
\label{eq:residual_fusion}
\end{equation}
where $\phi$ projects the direct vision features and the MLP learns a correction to $\mathbf v_i$ from the concatenated streams.

To condition spatial prediction on the instruction, we append a fixed action placeholder after the screenshot and instruction and use its final hidden state as the action query $\mathbf q$. This position can incorporate both inputs and supplies their joint context to the decoder. After projecting and reshaping the fused features into $X\in\mathbb{R}^{C\times h\times w}$, we apply query-dependent channel modulation~\citep{perez2018film}:
\begin{equation}
F_{:,r,c}=X_{:,r,c}\odot\bigl(1+\boldsymbol\gamma(\mathbf q)\bigr)+\boldsymbol\beta(\mathbf q),
\label{eq:conditioned_grid}
\end{equation}
where $\boldsymbol\gamma$ and $\boldsymbol\beta$ are learned scale and bias functions. Although shared across spatial positions, these modulation parameters act on position-dependent visual features, allowing the subsequent decoder to produce an instruction-specific mask.

\subsection{Sub-Token Action-Mask Prediction}
A single visual token can span both the target and surrounding controls. Assigning one score to that token cannot directly distinguish these locations. Seg2Click therefore predicts multiple scores within each token footprint. Let $s$ be the grid stride in processed pixels, so that $H'=sh$ and $W'=sw$. Each grid cell produces $s^2$ logits, one for each pixel offset within its $s\times s$ footprint.

A convolutional decoder aggregates neighboring conditioned features and outputs these logits. Pixel shuffle~\citep{shi2016espcn,wang2025mllmseg} then rearranges them into the dense field:
\begin{equation}
U=g(F)\in\mathbb{R}^{s^2\times h\times w},
\qquad L=\operatorname{PS}_s(U)\in\mathbb{R}^{H'\times W'},
\label{eq:pixel_shuffle_output}
\end{equation}
where $g$ consists of two spatial convolutional blocks followed by an output projection. Each output channel corresponds to a distinct within-cell offset. The learned decoder thus determines the sub-token scores, while pixel shuffle assigns them to their spatial locations. This differs from interpolating a token-level score map: the model learns separate outputs within each footprint, enabling finer localization while retaining the backbone's visual-token grid. Decoder configurations and the exact channel-to-pixel mapping are provided in Appendix A.

\subsection{Learning and Click Extraction.}
\label{sec:mask_to_click}
\textbf{Box-supervised learning.}
We rasterize the annotated target box into a binary mask $Y\in\{0,1\}^{H'\times W'}$ and train the prediction $P$ with
\begin{equation}
\mathcal L_{\mathrm{mask}}=
\tfrac12\mathcal L_{\mathrm{BCE}}(P,Y)+
\tfrac12\mathcal L_{\mathrm{Dice}}(P,Y).
\label{eq:batch_objective}
\end{equation}
Binary cross-entropy supervises pixel labels, while Dice loss measures overlap with the target region. This objective learns the spatial extent specified by existing box annotations without requiring contour labels. We optimize the fusion and decoder parameters together with LoRA adapters in the backbone~\citep{hu2021lora}, keeping the remaining parameters frozen. Appendix A specifies target rasterization, loss aggregation, and training settings.

\textbf{Deterministic click extraction.}
At inference, we threshold $P$ at $0.5$, select its largest four-connected foreground component, and return the centroid of its pixel centers as the click. Selecting one component avoids averaging across disconnected predictions. The normalized coordinate is mapped back to the original screenshot by scaling with its width and height. We use this fixed rule for all backbones and benchmarks; Algorithm~\ref{alg:seg2click} in Appendix~\ref{app:implementation} gives the full inference procedure. Each screenshot--instruction pair requires one mask-producing prediction, without coordinate-token generation, a verifier, or a zoom-in pass.

\begin{table}[!t]
\centering
\fontsize{8}{9.6}\selectfont
\setlength{\tabcolsep}{2.8pt}
\renewcommand{\arraystretch}{1.20}
\caption{\revision{Performance comparison on ScreenSpot-Pro, a benchmark of high-resolution professional interfaces. Dev, Creat., and Sci. denote Development, Creative, and Scientific, respectively.}}
\label{tab:screenspot_pro_main}
\begin{tabular}{@{}lllrrrrrrrr>{\columncolor{resultblue}}r@{}}
\toprule
& & & \multicolumn{6}{c}{Application domain} & \multicolumn{2}{c}{Target type} & \\
\cmidrule(lr){4-9}\cmidrule(lr){10-11}
Method & Backbone VLM & Venue / Year & Dev & Creat. & CAD & Sci. & Office & OS & Text & Icon & Avg \\
\midrule
Aria-UI & Aria-25.3B & ACL Findings 2025 & 8.4 & 14.7 & 6.1 & 18.1 & 16.1 & 2.6 & 17.1 & 2.0 & 11.3 \\
\midrule
CogAgent & CogVLM-17B & CVPR 2024 & 8.0 & 5.6 & 6.1 & 13.4 & 10.0 & 3.1 & 12.0 & 0.8 & 7.7 \\
\midrule
SeeClick & Qwen-VL-Chat & ACL 2024 & 0.3 & 0.6 & 1.9 & 2.0 & 0.9 & 1.5 & 1.8 & 0.0 & 1.1 \\
\midrule
ShowUI & Qwen2-VL-2B & CVPR 2025 & 9.4 & 5.3 & 1.9 & 10.6 & 13.5 & 6.6 & 10.8 & 2.6 & 7.7 \\
UGround-V1 & Qwen2-VL-2B & ICLR 2025 & 27.4 & 26.7 & 14.6 & 34.3 & 38.3 & 17.9 & -- & -- & 26.6 \\
UI-TARS & Qwen2-VL-2B & arXiv 2025 & 26.4 & 27.6 & 14.6 & 39.8 & 42.6 & 14.3 & 39.6 & 8.4 & 27.7 \\
\midrule
OS-Atlas & Qwen2-VL-7B & ICLR 2025 & 17.7 & 17.9 & 10.3 & 24.4 & 27.4 & 16.8 & 28.1 & 4.0 & 18.9 \\
Aguvis & Qwen2-VL-7B & ICML 2025 & 16.1 & 21.4 & 13.8 & 34.6 & 34.3 & 19.4 & -- & -- & 22.9 \\
UGround-V1 & Qwen2-VL-7B & ICLR 2025 & 28.1 & 31.7 & 14.6 & 39.0 & 49.6 & 24.5 & -- & -- & 31.1 \\
UI-TARS & Qwen2-VL-7B & arXiv 2025 & 36.1 & 32.8 & 18.0 & 50.0 & 53.5 & 24.5 & 47.8 & 16.2 & 35.7 \\
\midrule
Qwen2.5-VL & Qwen2.5-VL-3B & arXiv 2025 & 21.4 & 25.8 & 18.4 & 29.5 & 40.9 & 20.4 & -- & -- & 25.9 \\
Jedi & Qwen2.5-VL-3B & NeurIPS 2025 & 38.1 & 34.6 & 23.0 & 38.6 & 57.0 & 25.0 & -- & -- & 36.1 \\
GUI-Actor & Qwen2.5-VL-3B & NeurIPS 2025 & 39.8 & 36.7 & 34.1 & 49.6 & 61.3 & 35.2 & -- & -- & 42.2 \\
GUI-Actor + V. & Qwen2.5-VL-3B & NeurIPS 2025 & 43.8 & 37.8 & 43.3 & 48.4 & 63.5 & 42.9 & -- & -- & 45.9 \\
\rowcolor{resultblue}\textbf{Seg2Click} & Qwen2.5-VL-3B & Ours & 43.5 & 41.3 & 37.9 & \underline{53.1} & 67.8 & 41.8 & 62.7 & 21.5 & 47.0 \\
\midrule
Qwen2.5-VL & Qwen2.5-VL-7B & arXiv 2025 & 29.1 & 24.9 & 13.8 & 31.1 & 45.7 & 22.4 & -- & -- & 27.6 \\
Jedi & Qwen2.5-VL-7B & NeurIPS 2025 & 27.4 & 34.0 & 32.2 & 52.4 & 68.7 & 26.0 & -- & -- & 39.5 \\
GUI-Actor & Qwen2.5-VL-7B & NeurIPS 2025 & 38.1 & 41.4 & 38.3 & 50.8 & 63.0 & 38.8 & -- & -- & 44.6 \\
GUI-Actor + V. & Qwen2.5-VL-7B & NeurIPS 2025 & 43.1 & 41.9 & \underline{48.3} & 47.2 & 65.7 & 43.4 & -- & -- & 47.7 \\
\midrule
\rowcolor{resultblue}\textbf{Seg2Click} & Qwen3.5-0.8B & Ours & 40.1 & 41.3 & 35.6 & 46.9 & 64.3 & 40.3 & 61.5 & 16.4 & 44.3 \\
\rowcolor{resultblue}\textbf{Seg2Click} & Qwen3.5-2B & Ours & \revision{\textbf{47.2}} & 45.7 & 37.9 & \textbf{53.5} & 69.6 & 49.0 & \underline{65.2} & 25.0 & 49.8 \\
\rowcolor{resultblue}\textbf{Seg2Click} & Qwen3.5-4B & Ours & \revision{\underline{45.5}} & \underline{48.1} & 39.5 & 51.6 & \underline{70.0} & \revision{\underline{50.5}} & \underline{65.2} & \underline{26.0} & \revision{\underline{50.2}} \\
\rowcolor{resultblue}\textbf{Seg2Click} & Qwen3.5-9B & Ours & 44.1 & \textbf{53.7} & \textbf{59.0} & 48.0 & \textbf{77.4} & \textbf{53.1} & \textbf{68.9} & \textbf{33.1} & \textbf{55.2} \\
\bottomrule
\end{tabular}
\end{table}

\begin{table}[!t]
\centering
\fontsize{8}{9.6}\selectfont
\setlength{\tabcolsep}{2.8pt}
\renewcommand{\arraystretch}{1.20}
\caption{\revision{Performance comparison on ScreenSpot.}}
\label{tab:screenspot_main}
\begin{tabular}{@{}lllrrrrrr>{\columncolor{resultblue}}r@{}}
\toprule
& & & \multicolumn{2}{c}{Mobile} & \multicolumn{2}{c}{Desktop} & \multicolumn{2}{c}{Web} & \\
\cmidrule(lr){4-5}\cmidrule(lr){6-7}\cmidrule(lr){8-9}
Method & Backbone VLM & Venue / Year & Text & Icon & Text & Icon & Text & Icon & Avg \\
\midrule
CogAgent & CogVLM-17B & CVPR 2024 & 67.0 & 24.0 & 74.2 & 20.0 & 70.4 & 28.6 & 47.4 \\
\midrule
Magma & Llama-3-8B & CVPR 2025 & 60.4 & 58.5 & 75.3 & 52.9 & 69.1 & 52.0 & 60.3 \\
\midrule
SeeClick & Qwen-VL-Chat & ACL 2024 & 78.0 & 52.0 & 72.2 & 30.0 & 55.7 & 32.5 & 53.4 \\
\midrule
UGround-V1 & Qwen2-VL-2B & ICLR 2025 & 89.4 & 72.0 & 88.7 & 65.7 & 81.3 & 68.9 & 77.7 \\
UI-TARS & Qwen2-VL-2B & arXiv 2025 & 93.0 & 75.5 & 90.7 & 68.6 & 84.3 & 74.8 & 82.3 \\
GUI-Actor & Qwen2-VL-2B & NeurIPS 2025 & 93.0 & 79.9 & 88.1 & 78.6 & 90.9 & 84.0 & 86.5 \\
GUI-Actor + V. & Qwen2-VL-2B & NeurIPS 2025 & 93.8 & 81.2 & 89.7 & 80.7 & 91.3 & 80.6 & 86.9 \\
\midrule
Qwen2-VL & Qwen2-VL-7B & arXiv 2024 & 75.5 & 60.7 & 76.3 & 54.3 & 35.2 & 25.7 & 55.3 \\
Aguvis-G & Qwen2-VL-7B & ICML 2025 & 88.3 & 78.2 & 88.1 & 70.7 & 85.7 & 74.8 & 81.8 \\
OS-Atlas & Qwen2-VL-7B & ICLR 2025 & 93.0 & 72.9 & 91.8 & 62.9 & 90.9 & 74.3 & 82.5 \\
Aguvis & Qwen2-VL-7B & ICML 2025 & 95.6 & 77.7 & 93.8 & 67.1 & 88.3 & 75.2 & 84.4 \\
UGround-V1 & Qwen2-VL-7B & ICLR 2025 & 93.0 & 79.9 & 93.8 & 76.4 & 90.9 & 84.0 & 86.3 \\
UI-TARS & Qwen2-VL-7B & arXiv 2025 & 94.5 & \underline{85.2} & \underline{95.9} & 85.7 & 90.0 & 83.5 & 89.5 \\
GUI-Actor & Qwen2-VL-7B & NeurIPS 2025 & 94.9 & 82.1 & 91.8 & 80.0 & 91.3 & 85.4 & 88.3 \\
GUI-Actor + V. & Qwen2-VL-7B & NeurIPS 2025 & 96.0 & 83.0 & 93.8 & 82.1 & 92.2 & \underline{87.4} & 89.7 \\
\midrule
UI-TARS & Qwen2-VL-72B & arXiv 2025 & 94.9 & 82.5 & 89.7 & 88.6 & 88.7 & 85.0 & 88.4 \\
Aguvis & Qwen2-VL-72B & ICML 2025 & 94.5 & \underline{85.2} & 95.4 & 77.9 & 91.3 & 85.9 & 89.2 \\
UGround-V1 & Qwen2-VL-72B & ICLR 2025 & 94.1 & 83.4 & 94.9 & 85.7 & 90.4 & \textbf{87.9} & 89.4 \\
\midrule
\rowcolor{resultblue}\textbf{Seg2Click} & Qwen2.5-VL-3B & Ours & \underline{96.3} & 82.5 & \underline{95.9} & 82.1 & 92.6 & 81.1 & 89.1 \\
\midrule
\rowcolor{resultblue}\textbf{Seg2Click} & Qwen3.5-0.8B & Ours & 96.0 & 79.9 & 93.8 & 82.9 & 92.2 & 76.2 & 87.4 \\
\rowcolor{resultblue}\textbf{Seg2Click} & Qwen3.5-2B & Ours & 94.5 & 83.0 & 94.3 & 90.7 & \underline{93.0} & 82.5 & 89.8 \\
\rowcolor{resultblue}\textbf{Seg2Click} & Qwen3.5-4B & Ours & \textbf{96.7} & \underline{85.2} & \textbf{97.4} & \underline{91.4} & 90.9 & \underline{87.4} & \underline{91.6} \\
\rowcolor{resultblue}\textbf{Seg2Click} & Qwen3.5-9B & Ours & \underline{96.3} & \textbf{88.2} & \underline{95.9} & \textbf{92.9} & \textbf{93.5} & 85.0 & \textbf{92.1} \\
\bottomrule
\end{tabular}
\end{table}

\begin{table}[!t]
\centering
\fontsize{8}{9.6}\selectfont
\setlength{\tabcolsep}{2.8pt}
\renewcommand{\arraystretch}{1.20}
\caption{\revision{Performance comparison on ScreenSpot-v2. $\dagger$ denotes reported evaluations of official checkpoints.}}
\label{tab:screenspot_v2_main}
\begin{tabular}{@{}lllrrrrrr>{\columncolor{resultblue}}r@{}}
\toprule
& & & \multicolumn{2}{c}{Mobile} & \multicolumn{2}{c}{Desktop} & \multicolumn{2}{c}{Web} & \\
\cmidrule(lr){4-5}\cmidrule(lr){6-7}\cmidrule(lr){8-9}
Method & Backbone VLM & Venue / Year & Text & Icon & Text & Icon & Text & Icon & Avg \\
\midrule
OS-Atlas & InternVL2-4B & ICLR 2025 & 87.2 & 59.7 & 72.7 & 46.4 & 85.9 & 63.1 & 71.9 \\
\midrule
Magma & Llama-3-8B & CVPR 2025 & 62.8 & 53.4 & 80.0 & 57.9 & 67.5 & 47.3 & 61.5 \\
\midrule
SeeClick & Qwen-VL-Chat & ACL 2024 & 78.4 & 50.7 & 70.1 & 29.3 & 55.2 & 32.5 & 55.1 \\
\midrule
UI-TARS & Qwen2-VL-2B & arXiv 2025 & 95.2 & 79.1 & 90.7 & 68.6 & 87.2 & 78.3 & 84.7 \\
GUI-Actor & Qwen2-VL-2B & NeurIPS 2025 & 95.0 & 82.2 & 92.2 & 81.8 & 92.9 & 82.7 & 88.6 \\
GUI-Actor + V. & Qwen2-VL-2B & NeurIPS 2025 & 95.9 & 81.5 & 94.3 & 82.9 & 93.6 & 82.8 & 89.3 \\
\midrule
OS-Atlas & Qwen2-VL-7B & ICLR 2025 & 95.2 & 75.8 & 90.7 & 63.6 & 90.6 & 77.3 & 84.1 \\
Aguvis$^{\dagger}$ & Qwen2-VL-7B & ICML 2025 & 95.5 & 77.3 & 95.4 & 77.9 & 91.0 & 72.4 & 86.0 \\
UGround-V1$^{\dagger}$ & Qwen2-VL-7B & ICLR 2025 & 95.0 & 83.3 & 95.0 & 77.8 & 92.1 & 77.2 & 87.6 \\
UI-TARS & Qwen2-VL-7B & arXiv 2025 & 96.9 & 89.1 & 95.4 & 85.0 & 93.6 & 85.2 & 91.6 \\
GUI-Actor & Qwen2-VL-7B & NeurIPS 2025 & 96.5 & 84.3 & 91.7 & 84.1 & 93.9 & 82.3 & 89.5 \\
GUI-Actor + V. & Qwen2-VL-7B & NeurIPS 2025 & 97.2 & 84.8 & 94.3 & 85.0 & 94.0 & 85.2 & 90.9 \\
\midrule
UI-TARS & Qwen2-VL-72B & arXiv 2025 & 94.8 & 86.3 & 91.2 & 87.9 & 91.5 & \textbf{87.7} & 90.3 \\
\midrule
Qwen2.5-VL & Qwen2.5-VL-3B & arXiv 2025 & 93.4 & 73.5 & 88.1 & 58.6 & 88.0 & 71.4 & 80.9 \\
Jedi & Qwen2.5-VL-3B & NeurIPS 2025 & 96.6 & 81.5 & 96.9 & 78.6 & 88.5 & 83.7 & 88.6 \\
GUI-Actor & Qwen2.5-VL-3B & NeurIPS 2025 & 97.6 & 83.4 & 96.9 & 83.6 & 94.0 & 85.7 & 91.0 \\
GUI-Actor + V. & Qwen2.5-VL-3B & NeurIPS 2025 & 98.3 & 85.3 & 96.9 & 87.9 & 95.3 & 86.7 & 92.4 \\
\rowcolor{resultblue}\textbf{Seg2Click} & Qwen2.5-VL-3B & Ours & \underline{98.8} & 85.3 & 96.7 & 84.1 & 95.7 & 82.3 & 91.3 \\
\midrule
Qwen2.5-VL & Qwen2.5-VL-7B & arXiv 2025 & 97.6 & 87.2 & 90.2 & 74.2 & 93.2 & 81.3 & 88.8 \\
Jedi & Qwen2.5-VL-7B & NeurIPS 2025 & 96.9 & 87.2 & 95.9 & 87.9 & 94.4 & 84.2 & 91.7 \\
GUI-Actor & Qwen2.5-VL-7B & NeurIPS 2025 & 97.6 & 88.2 & 96.9 & 85.7 & 93.2 & 86.7 & 92.1 \\
GUI-Actor + V. & Qwen2.5-VL-7B & NeurIPS 2025 & 96.9 & \underline{89.6} & \underline{97.4} & 86.4 & 95.7 & 84.7 & 92.5 \\
\midrule
\rowcolor{resultblue}\textbf{Seg2Click} & Qwen3.5-0.8B & Ours & 97.7 & 81.7 & 94.4 & 87.3 & 93.6 & 81.9 & 89.9 \\
\rowcolor{resultblue}\textbf{Seg2Click} & Qwen3.5-2B & Ours & 95.7 & 86.9 & 95.0 & 92.1 & \underline{96.1} & 82.7 & 91.6 \\
\rowcolor{resultblue}\textbf{Seg2Click} & Qwen3.5-4B & Ours & 97.7 & 88.5 & \textbf{97.8} & \underline{92.9} & 93.9 & \underline{87.3} & \underline{93.1} \\
\rowcolor{resultblue}\textbf{Seg2Click} & Qwen3.5-9B & Ours & \textbf{99.2} & \textbf{90.1} & 96.7 & \textbf{95.2} & \textbf{96.4} & 86.5 & \textbf{94.1} \\
\bottomrule
\end{tabular}
\end{table}

\section{Experiments}
\label{sec:experiments}
We evaluate Seg2Click along three complementary axes: grounding accuracy across general and professional interfaces, the effectiveness of action masks relative to coordinate generation and patch scoring, and the contribution of the feature fusion and spatial decoder. Experiments cover five backbone configurations from two model families, using a common mask-to-click procedure throughout.

\subsection{Experimental Setup}
\label{sec:experimental_setup}
\textbf{Training.} We instantiate Seg2Click with Qwen3.5-0.8B, 2B, 4B, and 9B, and Qwen2.5-VL-3B~\citep{qwen2025qwen25vl,qwen2026qwen35,qwen2026qwen35four,qwen2026qwen35nine,qwen2026qwen35two}. All models use a 1,024-channel decoder and are trained for one epoch on the same grounding data as GUI-Actor, comprising approximately 1M screenshots from six public GUI datasets~\citep{wu2025guiactor}. We use LoRA with rank 16, an effective batch size of 32 across eight GPUs, and fused AdamW with a learning rate of $10^{-4}$ and cosine decay. Appendix A provides the complete training configuration and trainable parameter counts.

\textbf{Benchmarks and comparisons.} ScreenSpot and ScreenSpot-v2 each contain 1,272 examples spanning mobile, desktop, and web interfaces; ScreenSpot-Pro contains 1,581 examples from professional applications\citep{cheng2024seeclick,wu2025osatlas,li2025screenspotpro}. We report Hit@1: a prediction is correct when its single click lies inside the annotated target box, including its boundary. Overall accuracy is computed over all examples. Tables 1-3 report the performance of Seg2Click compared to existing advanced methods on the datasets ScreenSpot-Pro, ScreenSpot, and ScreenSpot-v2. Table 6 complements these comparisons with a matched-training coordinate baseline and a GUI-Actor reproduction on Qwen3.5-0.8B. Every Seg2Click evaluation uses the same mask threshold and largest-component centroid rule, with one mask-producing prediction per input. Measured inference costs are reported in Appendix~\ref{app:efficiency}.

\subsection{Benchmark Results and Analysis}
\label{sec:benchmark_results}

\paragraph{Overall grounding performance.}
Seg2Click-9B achieves 92.06\%, 94.10\%, and 55.22\% on ScreenSpot, ScreenSpot-v2, and ScreenSpot-Pro, respectively, leading the systems in Tables~1--3 by 2.36, 1.60, and 7.52 percentage points over the strongest published baselines. The largest margin occurs on ScreenSpot-Pro, highlighting its effectiveness in professional interfaces with small controls and dense layouts. All evaluations use the same decoder and deterministic click extraction, without a verifier or zoom-in pass.

\paragraph{Comparison on a shared backbone.}
With Qwen2.5-VL-3B, Seg2Click reaches 47.00\% on ScreenSpot-Pro, exceeding published GUI-Actor with the same base checkpoint by 4.80 points and its verifier-assisted variant by 1.10 points. Gains over GUI-Actor without a verifier span all six application domains, ranging from 3.5 to 6.6 points. Together with the 0.8B output-representation study in Section~4.3.3, these results support the effectiveness of the action-mask formulation for professional GUI grounding.

\paragraph{Localizing desktop icons and professional controls.}
Seg2Click-9B achieves 92.9\% and 95.2\% on desktop icons in ScreenSpot and ScreenSpot-v2, surpassing the strongest published baselines by 4.3 and 7.3 points. Seg2Click-4B also leads the corresponding desktop-text categories at 97.4\% and 97.8\%. On ScreenSpot-Pro, the Qwen3.5 variants collectively lead all six application domains: 9B leads Creative, CAD, Office, and OS, while 2B leads Development and Scientific. These results demonstrate broad applicability across target types and professional interfaces.

\paragraph{Scaling benefits icons and small targets.}
With the training recipe fixed, scaling Qwen3.5 from 0.8B to 9B improves ScreenSpot-Pro accuracy by 10.94 points, corresponding to 173 additional correct clicks. Gains are larger for icons than text (16.72 vs.\ 7.37 points). Accuracy increases by 13.90 points for targets occupying at most 0.01\% of the screenshot and by 11.77 points for those occupying 0.01\%--0.1\%, compared with 4.62 points for larger targets. These results indicate that backbone scaling particularly benefits icons and small targets; Appendix~\ref{sec:main_results} provides the full breakdown.

\subsection{Ablation Study}
\label{sec:ablations}

We use Qwen3.5-0.8B to examine feature sources, fusion strategies, and output representations. Feature and fusion comparisons retain the same mask objective and decoder width. The coordinate-generation comparison shares the initialization, training data, LoRA configuration, optimizer arguments, and one-epoch schedule with Seg2Click. Appendix~\ref{sec:decoder_capacity_ablation} additionally evaluates decoder capacity.

\subsubsection{Complementary Vision and Language-Model Features}

Table~4 shows that residual fusion achieves the highest accuracy among the evaluated feature configurations on all three benchmarks. It improves over final language-model image states by 0.55, 0.47, and 0.76 points on ScreenSpot, ScreenSpot-v2, and ScreenSpot-Pro. On ScreenSpot-Pro, the fused representation reaches 44.28\%, compared with 43.52\% for language-model states and 42.38\% for post-merge vision states. Improving over either stream alone supports their complementarity: direct vision features supply visual evidence, while final language-model states provide contextualized image representations. Combining the aligned streams gives the spatial decoder access to both sources when predicting the instruction-specific action region.

\begin{table}[!htbp]
\centering
\small
\caption{Feature sources at fixed Qwen3.5-0.8B and decoder width $C=1024$. \revision{Combining LLM and post-merge vision states improves all three benchmarks, reaching 44.28\% on ScreenSpot-Pro.} Accuracy is in percent ($\uparrow$). Best and second-best distinct values are bold and underlined, respectively; ties use the displayed precision.}
\label{tab:feature}
\begin{tabular}{@{}lrrr@{}}
\toprule
Feature source & SS & SS-v2 & SS-Pro \\
\midrule
Pre-merge only & 81.68 & 84.43 & 22.01 \\
Post-merge only & 86.32 & \underline{89.54} & 42.38 \\
LLM only, $C=1024$ & \underline{86.87} & 89.47 & \underline{43.52} \\
Residual concat--MLP & \textbf{87.42} & \textbf{89.94} & \textbf{44.28} \\
\bottomrule
\end{tabular}
\end{table}

The comparison between vision features also highlights the value of the backbone's learned spatial aggregation. Post-merge features reach 42.38\% on ScreenSpot-Pro, improving over pre-merge features by 20.37 points, or 322 correct clicks. This finding supports a two-stage design: use the backbone's aggregated representations to identify relevant visual content, then learn distinct spatial outputs within each token footprint. Seg2Click couples these informative features with dense decoding to recover the localization granularity needed for click selection.

\subsubsection{Residual Fusion and Decoder Capacity}

Table~5 compares residual concat--MLP fusion with a scalar-gated residual connection. Concat--MLP reaches 87.42\% and 89.94\% on ScreenSpot and ScreenSpot-v2, improving over scalar gating by 1.81 and 1.26 points. Both fusion strategies improve ScreenSpot-Pro over the unfused 43.52\% baseline, reaching 44.28\% with concat--MLP and 44.53\% with scalar gating. Concat--MLP provides gains over the unfused representation on every benchmark and the strongest results on the two standard benchmarks, motivating its use in the final architecture. Its joint transformation of the two feature streams allows the visual correction to depend on their combined content, while the residual path preserves the language-model representation.

\begin{table}[!htbp]
\centering
\small
\caption{Feature fusion at fixed Qwen3.5-0.8B and $C=1024$. \revision{Residual concat--MLP improves over unfused LLM states on all three benchmarks.} Accuracy is in percent ($\uparrow$). Best and second-best distinct values are bold and underlined, respectively; ties use the displayed precision.}
\label{tab:fusion}
\begin{tabular}{@{}lrrr@{}}
\toprule
Fusion & SS & SS-v2 & SS-Pro \\
\midrule
None (LLM only) & \underline{86.87} & \underline{89.47} & 43.52 \\
Scalar-gated residual & 85.61 & 88.68 & \textbf{44.53} \\
Residual concat--MLP & \textbf{87.42} & \textbf{89.94} & \underline{44.28} \\
\bottomrule
\end{tabular}
\end{table}

Decoder capacity interacts with feature fusion. With language-model image states alone, increasing the width from 128 to 1,024 channels improves the three benchmarks by 0.39, 0.16, and 1.39 points. With concat--MLP fusion, $C=256$ matches $C=1024$ on ScreenSpot-Pro, while $C=1024$ yields the highest ScreenSpot and ScreenSpot-v2 scores (Appendix~\ref{sec:decoder_capacity_ablation}) and remains the default configuration.

\subsubsection{Action Masks Improve Grounding Accuracy}
\label{sec:output_representation_ablation}

Table~6 directly evaluates the output representation on Qwen3.5-0.8B. Under matched training settings, Seg2Click improves over coordinate generation by 0.94, 1.02, and 1.52 points across the three benchmarks, adding 12, 13, and 24 correct clicks. The desktop subsets contribute 13 additional correct predictions on each standard benchmark, corresponding to gains of 3.89 and 4.25 points. On ScreenSpot-Pro, text-target accuracy increases from 58.85\% to 61.51\%. These results support learning the region of valid click locations before selecting the final point: box-derived supervision supplies a spatial target that is directly aligned with the box-hit evaluation criterion.

\begin{table}[!htbp]
\centering\small
\setlength{\tabcolsep}{5pt}
\caption{\revision{Seg2Click leads the compared output representations on all three benchmarks with Qwen3.5-0.8B.} GUI-Actor~\citep{wu2025guiactor} is reproduced on this backbone. \revision{Appendix~\ref{app:parameter_allocation} reports parameter counts.} Accuracy is in percent ($\uparrow$). Best and second-best distinct values are bold and underlined, respectively; ties use the displayed precision.}
\label{tab:output_representation}
\begin{tabular}{@{}llrrr@{}}
\toprule
Output representation & Backbone VLM & SS & SS-v2 & SS-Pro \\
\midrule
Direct coordinates & Qwen3.5-0.8B & 86.48 & 88.92 & \underline{42.76} \\
GUI-Actor & Qwen3.5-0.8B & \underline{87.26} & \underline{89.47} & 39.41 \\
Seg2Click & Qwen3.5-0.8B & \textbf{87.42} & \textbf{89.94} & \textbf{44.28} \\
\bottomrule
\end{tabular}
\end{table}

Relative to reproduced GUI-Actor on the same backbone, Seg2Click improves by 0.16, 0.47, and 4.87 points, with 77 additional correct predictions on ScreenSpot-Pro. Office, CAD, and Scientific applications contribute 20, 19, and 19 net additional hits, respectively, accounting for 58 of the 77-hit gain. This concentration across professional domains is consistent with the benefit of sub-token action prediction, which learns separate scores for locations within a visual-token footprint. Taken together, the coordinate and patch-scoring comparisons provide complementary evidence for Seg2Click's central design: represent valid click regions explicitly and learn their spatial structure at finer granularity than the visual-token grid.

\FloatBarrier
\section{\revision{Conclusion}}
\label{sec:conclusion}
\revision{Seg2Click formulates GUI grounding as box-supervised sub-token action-field prediction, learning fine-grained valid-click regions before deterministic point selection. Matched coordinate comparisons demonstrate accuracy gains across all three benchmarks, while the same-backbone patch-scoring comparison adds 77 correct ScreenSpot-Pro clicks. Evaluation across two model families reaches 55.22\% on ScreenSpot-Pro at 9B and 47.00\% at 3B. These results establish learned action masks as an effective output representation for precise GUI grounding.}

\revision{Building on these grounding results, future work will explore confidence-aware click selection and integration into interactive GUI agents, informed by the qualitative analysis in Appendix~\ref{app:qualitative}}.
\label{sec:conclusion_end}
\clearpage
\section*{AI Use Statement}
Generative AI tools assisted with language polishing and presentation refinement during manuscript writing. The authors reviewed the final manuscript and take responsibility for its content.

\section*{Ethics Statement}
This work evaluates grounding on existing GUI datasets. Screenshots can contain personal or proprietary content, and redistribution must respect source licenses and privacy constraints. Grounding supports accessibility and authorized automation. Responsible deployment requires user authorization and action-appropriate safeguards to prevent unintended interface actions.

\section*{Reproducibility Statement}
Section~\ref{sec:method} specifies the complete decoder and click rule, including mask-to-click decoding in Section~\ref{sec:mask_to_click}. Section~\ref{sec:experimental_setup} and Appendices~\ref{app:implementation} and~\ref{app:efficiency} document training settings, data counts, evaluation protocols, numerical precision, and inference settings.

\begingroup
\let\originalbibitem\bibitem
\renewcommand{\bibitem}[2][]{%
  \par\color{black}%
  \originalbibitem[#1]{#2}%
  \def\revisionbibkey{#2}%
  \def\revisiontargetkey{agentsea2024waveui}%
  \ifshowrevisions\ifx\revisionbibkey\revisiontargetkey\color{red}\fi\fi
  \ignorespaces
}
\bibliography{iclr2027_conference}

@inproceedings{cheng2024seeclick,
  title = {{SeeClick}: Harnessing {GUI} Grounding for Advanced Visual {GUI} Agents},
  author = {Cheng, Kanzhi and Sun, Qiushi and Chu, Yougang and Xu, Fangzhi and Li, Yantao and Zhang, Jianbing and Wu, Zhiyong},
  booktitle = {Proceedings of the 62nd Annual Meeting of the Association for Computational Linguistics (Volume 1: Long Papers)},
  pages = {9313--9332},
  year = {2024},
  publisher = {Association for Computational Linguistics},
  doi = {10.18653/v1/2024.acl-long.505},
  url = {https://aclanthology.org/2024.acl-long.505/}
}

@misc{agentsea2024waveui,
  title = {{Wave-UI}},
  author = {{AgentSea}},
  year = {2024},
  howpublished = {Hugging Face dataset},
  url = {https://huggingface.co/datasets/agentsea/wave-ui}
}

@inproceedings{gou2025uground,
  title = {Navigating the Digital World as Humans Do: Universal Visual Grounding for {GUI} Agents},
  author = {Gou, Boyu and Wang, Ruohan and Zheng, Boyuan and Xie, Yanan and Chang, Cheng and Shu, Yiheng and Sun, Huan and Su, Yu},
  booktitle = {International Conference on Learning Representations},
  year = {2025},
  url = {https://proceedings.iclr.cc/paper_files/paper/2025/hash/4ca0e369689dadb25a5345ba9755ad6f-Abstract-Conference.html}
}

@inproceedings{wu2025guiactor,
  title = {{GUI-Actor}: Coordinate-Free Visual Grounding for {GUI} Agents},
  author = {Wu, Qianhui and Cheng, Kanzhi and Yang, Rui and Zhang, Chaoyun and Yang, Jianwei and Jiang, Huiqiang and Mu, Jian and Peng, Baolin and Qiao, Bo and Tan, Reuben and Qin, Si and Liden, Lars and Lin, Qingwei and Zhang, Huan and Zhang, Tong and Zhang, Jianbing and Zhang, Dongmei and Gao, Jianfeng},
  booktitle = {Advances in Neural Information Processing Systems},
  volume = {38},
  doi = {10.52202/085713-0508},
  year = {2025},
  url = {https://proceedings.neurips.cc/paper_files/paper/2025/hash/16130af940e9dabb43c726119bd3b42e-Abstract-Conference.html}
}

@inproceedings{wu2025osatlas,
  title = {{OS-ATLAS}: A Foundation Action Model for Generalist {GUI} Agents},
  author = {Wu, Zhiyong and Wu, Zhenyu and Xu, Fangzhi and Wang, Yian and Sun, Qiushi and Jia, Chengyou and Cheng, Kanzhi and Ding, Zichen and Chen, Liheng and Liang, Paul Pu and Qiao, Yu},
  booktitle = {International Conference on Learning Representations},
  year = {2025},
  url = {https://proceedings.iclr.cc/paper_files/paper/2025/hash/0faa4bc5f522076947a030273629d4fe-Abstract-Conference.html}
}

@inproceedings{li2025screenspotpro,
  title = {{ScreenSpot-Pro}: {GUI} Grounding for Professional High-Resolution Computer Use},
  author = {Li, Kaixin and Meng, Ziyang and Lin, Hongzhan and Luo, Ziyang and Tian, Yuchen and Ma, Jing and Huang, Zhiyong and Chua, Tat-Seng},
  booktitle = {Proceedings of the 33rd ACM International Conference on Multimedia},
  pages = {8778--8786},
  publisher = {Association for Computing Machinery},
  doi = {10.1145/3746027.3755688},
  year = {2025},
  url = {https://doi.org/10.1145/3746027.3755688}
}

@article{wang2025mllmseg,
  title = {Unlocking the Potential of {MLLMs} in Referring Expression Segmentation via a Light-weight Mask Decoder},
  author = {Wang, Jingchao and Wu, Zhijian and Huang, Dingjiang and Zheng, Yefeng and Wang, Hong},
  journal = {arXiv preprint arXiv:2508.04107},
  year = {2025},
  url = {https://arxiv.org/abs/2508.04107}
}

@inproceedings{shi2016espcn,
  title = {Real-Time Single Image and Video Super-Resolution Using an Efficient Sub-Pixel Convolutional Neural Network},
  author = {Shi, Wenzhe and Caballero, Jose and Husz{\'a}r, Ferenc and Totz, Johannes and Aitken, Andrew P. and Bishop, Rob and Rueckert, Daniel and Wang, Zehan},
  booktitle = {Proceedings of the IEEE Conference on Computer Vision and Pattern Recognition},
  year = {2016},
  doi = {10.1109/CVPR.2016.207},
  url = {https://arxiv.org/abs/1609.05158}
}

@misc{qwen2026qwen35,
  title = {{Qwen3.5-0.8B}},
  author = {{Qwen Team}},
  year = {2026},
  howpublished = {Official model card},
  url = {https://huggingface.co/Qwen/Qwen3.5-0.8B}
}

@article{qin2025uitars,
  title = {{UI-TARS}: Pioneering Automated {GUI} Interaction with Native Agents},
  author = {Qin, Yujia and Ye, Yining and Fang, Junjie and Wang, Haoming and Liang, Shihao and Tian, Shizuo and Zhang, Junda and Li, Jiahao and Li, Yunxin and Huang, Shijue and others},
  journal = {arXiv preprint arXiv:2501.12326},
  year = {2025},
  url = {https://arxiv.org/abs/2501.12326}
}

@inproceedings{xu2025tag,
  title = {Attention-Driven {GUI} Grounding: Leveraging Pretrained Multimodal Large Language Models Without Fine-Tuning},
  author = {Xu, Hai-Ming and Chen, Qi and Wang, Lei and Liu, Lingqiao},
  booktitle = {Proceedings of the AAAI Conference on Artificial Intelligence},
  volume = {39},
  pages = {8851--8859},
  year = {2025},
  doi = {10.1609/aaai.v39i8.32957},
  url = {https://ojs.aaai.org/index.php/AAAI/article/view/32957}
}

@inproceedings{lai2024lisa,
  title = {{LISA}: Reasoning Segmentation via Large Language Model},
  author = {Lai, Xin and Tian, Zhuotao and Chen, Yukang and Li, Yanwei and Yuan, Yuhui and Liu, Shu and Jia, Jiaya},
  booktitle = {Proceedings of the IEEE/CVF Conference on Computer Vision and Pattern Recognition},
  pages = {9579--9589},
  year = {2024},
  url = {https://openaccess.thecvf.com/content/CVPR2024/html/Lai_LISA_Reasoning_Segmentation_via_Large_Language_Model_CVPR_2024_paper.html}
}

@inproceedings{ren2024pixellm,
  title = {{PixelLM}: Pixel Reasoning with Large Multimodal Model},
  author = {Ren, Zhongwei and Huang, Zhicheng and Wei, Yunchao and Zhao, Yao and Fu, Dongmei and Feng, Jiashi and Jin, Xiaojie},
  booktitle = {Proceedings of the IEEE/CVF Conference on Computer Vision and Pattern Recognition},
  pages = {26374--26383},
  year = {2024},
  url = {https://openaccess.thecvf.com/content/CVPR2024/html/Ren_PixelLM_Pixel_Reasoning_with_Large_Multimodal_Model_CVPR_2024_paper.html}
}

@inproceedings{perez2018film,
  title = {{FiLM}: Visual Reasoning with a General Conditioning Layer},
  author = {Perez, Ethan and Strub, Florian and de Vries, Harm and Dumoulin, Vincent and Courville, Aaron},
  booktitle = {Proceedings of the AAAI Conference on Artificial Intelligence},
  volume = {32},
  year = {2018},
  doi = {10.1609/aaai.v32i1.11671},
  url = {https://ojs.aaai.org/index.php/AAAI/article/view/11671}
}

@inproceedings{tian2021boxinst,
  title = {{BoxInst}: High-Performance Instance Segmentation With Box Annotations},
  author = {Tian, Zhi and Shen, Chunhua and Wang, Xinlong and Chen, Hao},
  booktitle = {Proceedings of the IEEE/CVF Conference on Computer Vision and Pattern Recognition},
  pages = {5443--5452},
  year = {2021},
  url = {https://openaccess.thecvf.com/content/CVPR2021/html/Tian_BoxInst_High-Performance_Instance_Segmentation_With_Box_Annotations_CVPR_2021_paper.html}
}

@inproceedings{hu2021lora,
  title = {{LoRA}: Low-Rank Adaptation of Large Language Models},
  author = {Hu, Edward J. and Shen, Yelong and Wallis, Phillip and Allen-Zhu, Zeyuan and Li, Yuanzhi and Wang, Shean and Wang, Lu and Chen, Weizhu},
  booktitle = {International Conference on Learning Representations},
  year = {2022},
  url = {https://openreview.net/forum?id=nZeVKeeFYf9}
}

@inproceedings{xu2024aguvis,
  title = {{Aguvis: Unified Pure Vision Agents for Autonomous GUI Interaction}},
  author = {Xu, Yiheng and Wang, Zekun and Wang, Junli and Lu, Dunjie and Xie, Tianbao and Saha, Amrita and Sahoo, Doyen and Yu, Tao and Xiong, Caiming},
  booktitle = {Proceedings of the 42nd International Conference on Machine Learning},
  pages = {69772--69805},
  year = {2025},
  series = {Proceedings of Machine Learning Research},
  volume = {267},
  publisher = {PMLR},
  url = {https://proceedings.mlr.press/v267/xu25ae.html}
}

@inproceedings{lin2024showui,
  title = {{ShowUI: One Vision-Language-Action Model for GUI Visual Agent}},
  author = {Lin, Kevin Qinghong and Li, Linjie and Gao, Difei and Yang, Zhengyuan and Wu, Shiwei and Bai, Zechen and Lei, Stan Weixian and Wang, Lijuan and Shou, Mike Zheng},
  booktitle = {Proceedings of the IEEE/CVF Conference on Computer Vision and Pattern Recognition},
  pages = {19498--19508},
  year = {2025},
  url = {https://openaccess.thecvf.com/content/CVPR2025/html/Lin_ShowUI_One_Vision-Language-Action_Model_for_GUI_Visual_Agent_CVPR_2025_paper.html}
}

@inproceedings{xie2025jedi,
  title = {Scaling Computer-Use Grounding via User Interface Decomposition and Synthesis},
  author = {Xie, Tianbao and Deng, Jiaqi and Li, Xiaochuan and Yang, Junlin and Wu, Haoyuan and Chen, Jixuan and Hu, Wenjing and Wang, Xinyuan and Xu, Yuhui and Wang, Zekun and Xu, Yiheng and Wang, Junli and Sahoo, Doyen and Yu, Tao and Xiong, Caiming},
  year = {2025},
  url = {https://proceedings.neurips.cc/paper_files/paper/2025/hash/22c868099177ee278eb7baccec649f35-Abstract-Datasets_and_Benchmarks_Track.html},
  booktitle = {Advances in Neural Information Processing Systems},
  volume = {38},
  doi = {10.52202/085713-0817},
}

@misc{qwen2026qwen35two,
  title = {{Qwen3.5-2B}},
  author = {{Qwen Team}},
  year = {2026},
  howpublished = {Official model card},
  url = {https://huggingface.co/Qwen/Qwen3.5-2B}
}

@article{qwen2025qwen25vl,
  title = {{Qwen2.5-VL} Technical Report},
  author = {Bai, Shuai and Chen, Keqin and Liu, Xuejing and Wang, Jialin and Ge, Wenbin and Song, Sibo and Dang, Kai and Wang, Peng and Wang, Shijie and Tang, Jun and others},
  journal = {arXiv preprint arXiv:2502.13923},
  year = {2025},
  url = {https://arxiv.org/abs/2502.13923}
}

@misc{qwen2026qwen35four,
  title = {{Qwen3.5-4B}},
  author = {{Qwen Team}},
  year = {2026},
  howpublished = {Official model card},
  url = {https://huggingface.co/Qwen/Qwen3.5-4B}
}

@misc{qwen2026qwen35nine,
  title = {{Qwen3.5-9B}},
  author = {{Qwen Team}},
  year = {2026},
  howpublished = {Official model card},
  url = {https://huggingface.co/Qwen/Qwen3.5-9B}
}

@inproceedings{rawles2025androidworld,
  title = {{AndroidWorld}: A Dynamic Benchmarking Environment for Autonomous Agents},
  author = {Rawles, Chris and Clinckemaillie, Sarah and Chang, Yifan and Waltz, Jonathan and Lau, Gabrielle and Fair, Marybeth and Li, Alice and Bishop, William and Li, Wei and Campbell-Ajala, Folawiyo and Toyama, Daniel and Berry, Robert and Tyamagundlu, Divya and Lillicrap, Timothy and Riva, Oriana},
  booktitle = {International Conference on Learning Representations},
  pages = {406--441},
  year = {2025},
  url = {https://proceedings.iclr.cc/paper_files/paper/2025/hash/01a83bc2f2732a58e6aa731e659e7101-Abstract-Conference.html},
}

@inproceedings{zhou2024webarena,
  title = {{WebArena}: A Realistic Web Environment for Building Autonomous Agents},
  author = {Zhou, Shuyan and Xu, Frank F. and Zhu, Hao and Zhou, Xuhui and Lo, Robert and Sridhar, Abishek and Cheng, Xianyi and Ou, Tianyue and Bisk, Yonatan and Fried, Daniel and Alon, Uri and Neubig, Graham},
  booktitle = {International Conference on Learning Representations},
  pages = {15585--15606},
  year = {2024},
  url = {https://proceedings.iclr.cc/paper_files/paper/2024/hash/4410c0711e9154a7a2d26f9b3816d1ef-Abstract-Conference.html},
}

@inproceedings{deng2023mind2web,
  title = {{Mind2Web}: Towards a Generalist Agent for the Web},
  author = {Deng, Xiang and Gu, Yu and Zheng, Boyuan and Chen, Shijie and Stevens, Sam and Wang, Boshi and Sun, Huan and Su, Yu},
  booktitle = {Advances in Neural Information Processing Systems},
  volume = {36},
  pages = {28091--28114},
  year = {2023},
  doi = {10.52202/075280-1220},
  url = {https://papers.neurips.cc/paper_files/paper/2023/hash/5950bf290a1570ea401bf98882128160-Abstract-Datasets_and_Benchmarks.html},
}

@inproceedings{xie2024osworld,
  title = {{OSWorld}: Benchmarking Multimodal Agents for Open-Ended Tasks in Real Computer Environments},
  author = {Xie, Tianbao and Zhang, Danyang and Chen, Jixuan and Li, Xiaochuan and Zhao, Siheng and Cao, Ruisheng and Hua, Toh Jing and Cheng, Zhoujun and Shin, Dongchan and Lei, Fangyu and Liu, Yitao and Xu, Yiheng and Zhou, Shuyan and Savarese, Silvio and Xiong, Caiming and Zhong, Victor and Yu, Tao},
  booktitle = {Advances in Neural Information Processing Systems},
  volume = {37},
  pages = {52040--52094},
  year = {2024},
  doi = {10.52202/079017-1650},
  url = {https://proceedings.neurips.cc/paper_files/paper/2024/hash/5d413e48f84dc61244b6be550f1cd8f5-Abstract-Datasets_and_Benchmarks_Track.html},
}

@inproceedings{he2024webvoyager,
  title = {{WebVoyager}: Building an End-to-End Web Agent with Large Multimodal Models},
  author = {He, Hongliang and Yao, Wenlin and Ma, Kaixin and Yu, Wenhao and Dai, Yong and Zhang, Hongming and Lan, Zhenzhong and Yu, Dong},
  booktitle = {Proceedings of the 62nd Annual Meeting of the Association for Computational Linguistics (Volume 1: Long Papers)},
  pages = {6864--6890},
  year = {2024},
  publisher = {Association for Computational Linguistics},
  doi = {10.18653/v1/2024.acl-long.371},
  url = {https://aclanthology.org/2024.acl-long.371/},
}

@inproceedings{chen2025guicourse,
  title = {{GUICourse}: From General Vision Language Model to Versatile {GUI} Agent},
  author = {Chen, Wentong and Cui, Junbo and Hu, Jinyi and Qin, Yujia and Fang, Junjie and Zhao, Yue and Wang, Chongyi and Liu, Jun and Chen, Guirong and Huo, Yupeng and Yao, Yuan and Lin, Yankai and Liu, Zhiyuan and Sun, Maosong},
  booktitle = {Proceedings of the 63rd Annual Meeting of the Association for Computational Linguistics (Volume 1: Long Papers)},
  pages = {21936--21959},
  publisher = {Association for Computational Linguistics},
  year = {2025},
  doi = {10.18653/v1/2025.acl-long.1065},
  url = {https://aclanthology.org/2025.acl-long.1065/}
}

@inproceedings{chai2025amex,
  title = {{AMEX}: Android Multi-annotation Expo Dataset for Mobile {GUI} Agents},
  author = {Chai, Yuxiang and Huang, Siyuan and Niu, Yazhe and Xiao, Han and Liu, Liang and Wang, Guozhi and Zhang, Dingyu and Ren, Shuai and Li, Hongsheng},
  booktitle = {Findings of the Association for Computational Linguistics: ACL 2025},
  pages = {2138--2156},
  publisher = {Association for Computational Linguistics},
  year = {2025},
  doi = {10.18653/v1/2025.findings-acl.110},
  url = {https://aclanthology.org/2025.findings-acl.110/}
}

@inproceedings{li2024androidcontrol,
  title = {On the Effects of Data Scale on {UI} Control Agents},
  author = {Li, Wei and Bishop, William and Li, Alice and Rawles, Chris and Campbell-Ajala, Folawiyo and Tyamagundlu, Divya and Riva, Oriana},
  booktitle = {Advances in Neural Information Processing Systems},
  volume = {37},
  pages = {92130--92154},
  publisher = {Curran Associates, Inc.},
  year = {2024},
  doi = {10.52202/079017-2925},
  url = {https://papers.nips.cc/paper_files/paper/2024/hash/a79f3ef3b445fd4659f44648f7ea8ffd-Abstract-Datasets_and_Benchmarks_Track.html}
}

@inproceedings{you2024ferretui,
  title = {{Ferret-UI}: Grounded Mobile {UI} Understanding with Multimodal {LLMs}},
  author = {You, Keen and Zhang, Haotian and Schoop, Eldon and Weers, Floris and Swearngin, Amanda and Nichols, Jeffrey and Yang, Yinfei and Gan, Zhe},
  booktitle = {Computer Vision -- ECCV 2024},
  pages = {240--255},
  publisher = {Springer},
  year = {2024},
  doi = {10.1007/978-3-031-73039-9_14},
  url = {https://www.ecva.net/papers/eccv_2024/papers_ECCV/papers/08095.pdf}
}

@inproceedings{bai2021uibert,
  title = {{UIBert}: Learning Generic Multimodal Representations for {UI} Understanding},
  author = {Bai, Chongyang and Zang, Xiaoxue and Xu, Ying and Sunkara, Srinivas and Rastogi, Abhinav and Chen, Jindong and {Ag{\"u}era y Arcas}, Blaise},
  booktitle = {Proceedings of the Thirtieth International Joint Conference on Artificial Intelligence},
  pages = {1705--1712},
  publisher = {International Joint Conferences on Artificial Intelligence Organization},
  year = {2021},
  doi = {10.24963/ijcai.2021/235},
  url = {https://www.ijcai.org/proceedings/2021/235}
}

@inproceedings{yang2025ariaui,
  title = {{Aria-UI}: Visual Grounding for {GUI} Instructions},
  author = {Yang, Yuhao and Wang, Yue and Li, Dongxu and Luo, Ziyang and Chen, Bei and Huang, Chao and Li, Junnan},
  booktitle = {Findings of the Association for Computational Linguistics: ACL 2025},
  year = {2025},
  pages = {22418--22433},
  publisher = {Association for Computational Linguistics},
  doi = {10.18653/v1/2025.findings-acl.1152},
  url = {https://aclanthology.org/2025.findings-acl.1152/}
}

@inproceedings{pawlowski2025tinyclick,
  title = {{TinyClick}: Single-Turn Agent for Empowering {GUI} Automation},
  author = {Pawlowski, Pawel and Zawistowski, Krystian and Lapacz, Wojciech and Wiacek, Adam and Skorupa, Marcin and Postansque, Sebastien and Hoscilowicz, Jakub},
  booktitle = {Interspeech 2025},
  year = {2025},
  pages = {3035--3039},
  doi = {10.21437/Interspeech.2025-176},
  url = {https://www.isca-archive.org/interspeech_2025/pawlowski25_interspeech.html}
}

@article{chen2023shikra,
  title = {{Shikra}: Unleashing Multimodal {LLM}'s Referential Dialogue Magic},
  author = {Chen, Keqin and Zhang, Zhao and Zeng, Weili and Zhang, Richong and Zhu, Feng and Zhao, Rui},
  journal = {arXiv preprint arXiv:2306.15195},
  year = {2023},
  doi = {10.48550/arXiv.2306.15195},
  url = {https://arxiv.org/abs/2306.15195}
}

@article{lu2024omniparser,
  title = {{OmniParser} for Pure Vision Based {GUI} Agent},
  author = {Lu, Yadong and Yang, Jianwei and Shen, Yelong and Awadallah, Ahmed},
  journal = {arXiv preprint arXiv:2408.00203},
  year = {2024},
  doi = {10.48550/arXiv.2408.00203},
  url = {https://arxiv.org/abs/2408.00203}
}
\endgroup
\bibliographystyle{iclr2027_conference}
\clearpage
\appendix
\setlength{\textfloatsep}{12pt plus 2pt minus 2pt}
\setlength{\floatsep}{10pt plus 2pt minus 2pt}
\setlength{\intextsep}{10pt plus 2pt minus 2pt}
\setlength{\abovecaptionskip}{6pt}
\section{Training and Inference Details}
\label{app:implementation}

All five main Seg2Click models are initialized from official pretrained checkpoints and trained for one epoch using LoRA with rank 16, alpha 32, dropout 0.05, and all-linear targeting. The backbone operates in BF16, while the fusion and mask decoder operate in FP32. All models use a 1,024-channel spatial decoder. Qwen3.5 uses a patch size of 16 pixels and a pixel-shuffle factor of 32, whereas Qwen2.5-VL uses a patch size of 14 pixels and a pixel-shuffle factor of 28. Query conditioning, BCE/Dice loss weights, and centroid-based click extraction are identical across runs. The Qwen2.5-VL run additionally uses FlashAttention-2 in both the vision and language modules, together with gradient checkpointing in the language blocks. Each model is evaluated using its final checkpoint after one epoch. Table~\ref{tab:config} summarizes the model configurations.

We use the same training data as GUI-Actor, comprising the six sources listed in Table~\ref{tab:data_sources}. GUI-Env and GUI-Act originate from GUICourse~\citep{chen2025guicourse}, while AMEX and AndroidControl provide mobile-interface annotations and demonstrations~\citep{chai2025amex,li2024androidcontrol}. For Wave-UI~\citep{agentsea2024waveui}, we use the processed subset of 52,529 records distributed with GUI-Actor. According to GUI-Actor's data documentation, this released subset excludes samples overlapping the test set.\footnote{\url{https://huggingface.co/datasets/cckevinn/GUI-Actor-Data}} Wave-UI remains subject to the license terms of its constituent sources.\footnote{\url{https://huggingface.co/datasets/agentsea/wave-ui/blob/main/LICENSE.md}} Across all five runs, we retain at most ten non-system messages per record and compute the training loss by first averaging over target queries within each record and then averaging over records, giving each record equal weight.

\begin{table}[htbp]
\centering\fontsize{8.5}{10}\selectfont
\setlength{\tabcolsep}{3pt}
\caption{\revision{Training configurations across backbones. Parameter columns report LoRA adapters, the fusion/mask head, and their total; all runs use eight GPUs, effective batch 32 and 39,226 updates.}}
\label{tab:config}
\begin{tabular}{lrrrrrr}
\toprule
Backbone & Hidden $d$ & Batch/GPU & Accum. & LoRA params & \revision{Head params} & \revision{Total params}\\
\midrule
Qwen3.5-0.8B & 1024 & 4 & 1 & 13,345,792 & 27,286,528 & \revision{40,632,320}\\
Qwen3.5-2B & 2048 & 2 & 2 & 23,340,032 & 43,028,480 & \revision{66,368,512}\\
Qwen2.5-VL-3B & 2048 & 1 & 4 & 41,363,456 & 42,782,480 & \revision{84,145,936}\\
Qwen3.5-4B & 2560 & 1 & 4 & 38,993,920 & 54,045,184 & \revision{93,039,104}\\
Qwen3.5-9B & 4096 & 1 & 4 & 51,265,024 & 99,678,208 & \revision{150,943,232}\\
\bottomrule
\end{tabular}
\end{table}

\begin{table}[htbp]
\centering\small
\caption{The shared training mixture contains 1,255,208 records from six sources. Each training record can pair a screenshot with one or more queries.}
\label{tab:data_sources}
\begin{tabular}{lr}
\toprule Source & Records\\\midrule
UGround & 756,451\\ GUI-Env & 262,445\\ GUI-Act & 42,545\\
AMEX & 101,537\\ AndroidControl & 39,701\\ Wave-UI & 52,529\\
\midrule Total & 1,255,208\\\bottomrule
\end{tabular}
\end{table}

\paragraph{Optimization and numerical precision.}
The per-device batches are 4/2/1/1/1 and accumulation factors are 1/2/4/4/4 for Qwen3.5-0.8B, Qwen3.5-2B, Qwen2.5-VL-3B, Qwen3.5-4B, and Qwen3.5-9B, respectively.
We use zero weight decay, no warmup, a sequence limit of 24,576, and processor pixel bounds of 3,136 and 5,720,064.
LoRA targets linear layers in both the vision and language blocks while base weights remain frozen.
Decoder parameters are stored in FP32; the spatial mask head and its losses also execute in FP32, with a BF16 backbone.
New action-token embeddings use mean initialization and remain frozen during LoRA training.

\paragraph{\revision{Trainable parameters and grounding performance.}}
\label{app:parameter_allocation}
\revision{Table~\ref{tab:parameter_allocation} compares adapter and head capacity alongside grounding accuracy for the 0.8B models. The coordinate comparison matches backbone initialization, training data, adapter rank, and optimization. The head count includes learned feature fusion where enabled.}

\begin{table}[htbp]
\centering\fontsize{8.5}{10}\selectfont
\setlength{\tabcolsep}{3pt}
\caption{Trainable parameters and grounding accuracy for Qwen3.5-0.8B output architectures. Head counts include fusion where enabled. LLM-only rows use no direct vision fusion. Accuracy is in percent ($\uparrow$); best and second-best distinct available values are bold and underlined.}
\label{tab:parameter_allocation}
\begin{tabular}{@{}lrrrrrr@{}}
\toprule
& \multicolumn{3}{c}{Trainable parameters} & \multicolumn{3}{c}{Accuracy (\%)}\\
\cmidrule(lr){2-4}\cmidrule(l){5-7}
Configuration & LoRA & Head & Total & SS & SS-v2 & SS-Pro\\
\midrule
Direct coordinates & 13,345,792 & 0 & 13,345,792 & 86.48 & 88.92 & 42.76\\
GUI-Actor & 13,345,792 & 8,398,848 & 21,744,640 & \underline{87.26} & {89.47} & 39.41\\
LLM only, $C=128$ & 13,345,792 & 825,472 & 14,171,264 & 86.48 & 89.31 & 42.13\\
LLM only, $C=1024$ & 13,345,792 & 23,083,008 & 36,428,800 & 86.87 & {89.47} & {43.52}\\
\midrule
Seg2Click, fused $C=256$ & 13,345,792 & 6,439,168 & 19,784,960 & {86.56} & {\underline{89.86}} & {\textbf{44.28}}\\
Seg2Click, fused $C=512$ & 13,345,792 & 11,028,992 & 24,374,784 & {87.19} & 89.54 & \underline{44.09}\\
Seg2Click, fused $C=1024$ & 13,345,792 & 27,286,528 & 40,632,320 & \textbf{87.42} & \textbf{89.94} & \textbf{44.28}\\
\bottomrule
\end{tabular}
\end{table}

{The fused $C=256$ model uses 9.0\% fewer trainable parameters than GUI-Actor (19.78M versus 21.74M), with ScreenSpot-Pro accuracies of 44.28\% versus 39.41\%. Appendix~\ref{sec:decoder_capacity_ablation} analyzes the decoder-width trade-off.} \revision{These results describe the accuracy--capacity trade-off of the evaluated architectures; their differing parameter budgets do not isolate output representation from added capacity.}

\paragraph{Spatial decoder implementation.}
\label{app:decoder_details}
The two $3\times3$ convolutional blocks use stride one, padding one, group normalization with 32 groups, and GELU.
For grid position $(r,c)$ and within-cell offset $(a,b)$, pixel shuffle in Eq.~\eqref{eq:pixel_shuffle_output} uses
\begin{equation}
L_j[sr+a,sc+b]=\mathbf{U}_j[a\,s+b,r,c],
\qquad 0\leq a,b<s.
\label{eq:pixel_shuffle_mapping}
\end{equation}
Equation~\eqref{eq:pixel_shuffle_mapping} assigns the $s^2$ channels to distinct processed-image pixels without interpolation.

\paragraph{Target rasterization and loss definitions.}
For normalized target box $b_j$, the cell-overlap rule is
\begin{equation}
\begin{aligned}
R_{yx}&=\left[\frac{x}{W'},\frac{x+1}{W'}\right)
\times\left[\frac{y}{H'},\frac{y+1}{H'}\right),\\
Y_j(y,x)&=\mathbf{1}\!\left[\operatorname{area}(R_{yx}\cap b_j)>0\right].
\end{aligned}
\label{eq:box_target}
\end{equation}
Equation~\eqref{eq:box_target} defines the box-derived action target, which directly supervises the valid click region.
\revision{Training boxes must have finite coordinates in $[0,1]$ with $x_{\min}<x_{\max}$ and $y_{\min}<y_{\max}$; output grid dimensions must be positive. The positive-area overlap rule then yields at least one positive processed-pixel cell, including for subpixel boxes. Invalid boxes and anomalous empty rasterizations raise an error. This training-target check is separate from the inference-time empty-prediction fallback below.}
We define the losses used in Eq.~\eqref{eq:batch_objective} as
\begin{equation}
\begin{aligned}
\mathcal{L}_{\mathrm{BCE}}^{(j)}
&=-\frac{1}{H'W'}\sum_{y,x}
\left[Y_j(y,x)\log P_j(y,x)+(1-Y_j(y,x))\log(1-P_j(y,x))\right],\\
\mathcal{L}_{\mathrm{Dice}}^{(j)}
&=1-\frac{2\sum_{y,x}P_j(y,x)Y_j(y,x)+\epsilon}
{\sum_{y,x}P_j(y,x)+\sum_{y,x}Y_j(y,x)+\epsilon},
\qquad \epsilon=1.
\end{aligned}
\label{eq:mask_losses}
\end{equation}
In Eq.~\eqref{eq:mask_losses}, binary cross-entropy is implemented on logits for numerical stability.

\paragraph{Empty-mask fallback.}
When thresholding produces no foreground component, we compute the probability-weighted centroid
\begin{equation}
\hat{\mathbf{c}}_j
=\frac{\sum_{y,x}P_j(y,x)\mathbf{c}_{yx}}
{\sum_{y,x}P_j(y,x)}.
\label{eq:centroid_fallback}
\end{equation}
\revision{Here, $\mathbf c_{yx}=((x+\tfrac12)/W',(y+\tfrac12)/H')$ denotes the normalized pixel center.}
Equation~\eqref{eq:centroid_fallback} weights every pixel center by its predicted probability; a zero probability sum returns $(0.5,0.5)$.
{In the batch-size-one Qwen2.5-VL-3B evaluations in Appendix~\ref{app:efficiency}, empty-mask fallback occurs in 0/1,272 ScreenSpot examples, 0/1,272 ScreenSpot-v2 examples, and 1/1,581 ScreenSpot-Pro examples. The single fallback uses the probability-weighted centroid.}

\paragraph{Action prompt.}
The screenshot and instruction follow the grounding system message.
The fixed assistant suffix in the repository's action template is \nolinkurl{pyautogui.click(<|pointer_start|><|pointer_pad|><|pointer_end|>)}.
The \texttt{<|pointer\_pad|>} state is the action query.
At inference there is one instruction per example and one click per prediction, even when the numerical batch contains two examples.

\paragraph{Coordinate-generation baseline.}
\label{app:coordinate_protocol}
The matched training comparison appears in Section~\ref{sec:output_representation_ablation}.
\revision{The coordinate baseline preserves the source action coordinates supplied in the processed training records, already normalized to $[0,1]^2$ against the original screenshot. It formats four-decimal \nolinkurl{pyautogui.click(x=..., y=...)} targets using \texttt{ROUND\_HALF\_UP}, without recomputing points from box centers.}
It trains the adapters with cross-entropy averaged over supervised assistant tokens, including the backbone's native end-of-turn token \texttt{<|im\_end|>}, and uses no spatial head.
At inference, greedy decoding generates at most 64 tokens; a strict parser accepts finite in-range coordinates without clipping or repair.
\revision{All 4,125 coordinate predictions are valid and terminate with the native end-of-turn token.}

\begin{algorithm}[htbp]
\caption{Seg2Click inference for one screenshot and instruction}
\label{alg:seg2click}
\begin{tabular}{@{}r p{0.91\linewidth}@{}}
& \textbf{Input:} screenshot $I$ of size $H\times W$, instruction $u$, trained backbone and head.\\
& \textbf{Output:} one original-image click $(W\hat x,H\hat y)$.\\
1 & Preprocess $I'=\Pi(I)$; append the fixed action placeholder after $u$.\\
2 & Run the backbone; extract aligned $\{\mathbf z_i,\mathbf v_i\}$ and action query $\mathbf q$.\\
3 & Fuse features by Eq.~\eqref{eq:residual_fusion}; condition the grid by Eq.~\eqref{eq:conditioned_grid}.\\
4 & Apply Eq.~\eqref{eq:pixel_shuffle_output} to obtain $L\in\mathbb R^{H'\times W'}$; set $P=\sigma(L)$.\\
5 & Find four-connected components of $P\geq0.5$ and select the largest; break ties by smallest row-major index.\\
6 & If a component exists, use its unweighted centroid of normalized pixel centers; otherwise use Eq.~\eqref{eq:centroid_fallback}, returning $(0.5,0.5)$ for a zero sum.\\
7 & Return $(W\hat x,H\hat y)$.\\
\end{tabular}
\end{algorithm}

\begin{table}[htbp]
\centering\small
\caption{Notation links the backbone grid to the dense action output.}
\label{tab:notation}
\begin{tabular}{@{}ll@{}}
\toprule Symbol & Meaning\\\midrule
$H,W$; $H',W'$ & Original and processed screenshot dimensions\\
$p,m,s=pm$ & Patch size, merge factor, processed-pixel stride\\
$h,w,N=hw$ & Post-merge grid dimensions and token count\\
$d,C$ & Backbone hidden size and decoder width\\
$\mathbf z_i,\mathbf v_i,\mathbf f_i$ & Direct vision, final image, and fused features\\
$\mathbf q_j$ & Final action-query hidden state for query $j$\\
$L_j,P_j,Y_j$ & Dense logits, probabilities, and box-derived target\\
$\mathcal C^*,\hat{\mathbf c}_j$ & Selected component and normalized click\\
\bottomrule
\end{tabular}
\end{table}
Table~\ref{tab:notation} collects the symbols used in the method. Algorithm~\ref{alg:seg2click} fixes every inference choice independently of benchmark or backbone.

\FloatBarrier
\section{Additional Experimental Analysis}
\label{app:additional_experiments}
This appendix reports the backbone, target-size, and decoder-capacity analyses complementing the main comparisons in Section~\ref{sec:benchmark_results} and the ablations in Section~\ref{sec:ablations}.

\subsection{Grounding Performance and Backbone Scaling}
\label{sec:main_results}
\paragraph{Target-type and target-area analysis.}
\label{sec:fine_grained_analysis}
\label{app:analysis}
{Table~\ref{tab:fine_grained} reports the ScreenSpot-Pro breakdown for 977 text and 604 icon targets. Target area is $a=(x_{\max}-x_{\min})(y_{\max}-y_{\min})$, using box coordinates normalized by the original screenshot dimensions. The groups $a\leq0.01\%$, $0.01\%<a\leq0.1\%$, and $a>0.1\%$ contain 259, 1,062, and 260 targets, respectively.}

\begin{table}[!htbp]
\centering
\fontsize{8}{9.6}\selectfont
\setlength{\tabcolsep}{4.2pt}
\caption{Seg2Click accuracy on ScreenSpot-Pro (\%, $\uparrow$) by target type and normalized target-box area across backbones. Best and second-best distinct values are bold and underlined, respectively; ties use the displayed precision.}
\label{tab:fine_grained}
\begin{tabular}{@{}lrrrrrr@{}}
\toprule
& \multicolumn{3}{c}{Target type} & \multicolumn{3}{c}{Normalized target area $a$} \\
\cmidrule(lr){2-4}\cmidrule(l){5-7}
Backbone VLM & Text & Icon & Overall & $a\leq0.01\%$ & $0.01\%<a\leq0.1\%$ & $a>0.1\%$ \\
\midrule
Qwen3.5-0.8B & 61.51 & 16.39 & 44.28 & 9.27 & 46.33 & 70.77 \\
Qwen3.5-2B & \underline{65.20} & 25.00 & 49.84 & \underline{17.37} & 51.51 & \textbf{75.38} \\
Qwen2.5-VL-3B & 62.74 & 21.52 & 47.00 & 11.20 & 49.81 & 71.15 \\
Qwen3.5-4B & \underline{65.20} & \underline{25.99} & \underline{50.22} & 15.06 & \underline{53.48} & \underline{71.92} \\
Qwen3.5-9B & \textbf{68.88} & \textbf{33.11} & \textbf{55.22} & \textbf{23.17} & \textbf{58.10} & \textbf{75.38} \\
\bottomrule
\end{tabular}
\end{table}

{Scaling from 2B to 4B redistributes gains across target sizes: the intermediate-area group adds 21 correct predictions, while the smallest and largest groups lose 6 and 9, respectively, leaving a net gain of six. Text accuracy remains at 65.20\%. Thus, a higher overall score does not imply improvement in every target group.}

{The intermediate-area group contributes 125 of the 173 additional correct predictions from 0.8B to 9B, versus 36 and 12 from the smallest and largest groups. As the most populous group, its contribution reflects both its frequency and improved accuracy.}

\subsection{Does additional decoder capacity improve grounding?}
\label{sec:decoder_capacity_ablation}
\begin{table}[!htbp]
\centering
\small
\caption{{Decoder-width ablation on Qwen3.5-0.8B with residual concat--MLP fusion. Accuracy is in percent ($\uparrow$). Best and second-best distinct values are bold and underlined, respectively; ties use the displayed precision.}}
\label{tab:capacity}
\begin{tabular}{@{}lrrr@{}}
\toprule
Decoder channels & SS & SS-v2 & SS-Pro \\
\midrule
{256} & {86.56} & {\underline{89.86}} & {\textbf{44.28}} \\
{512} & {\underline{87.19}} & {89.54} & {\underline{44.09}} \\
{1024} & {\textbf{87.42}} & {\textbf{89.94}} & {\textbf{44.28}} \\
\bottomrule
\end{tabular}
\end{table}

{Table~\ref{tab:capacity} compares decoder widths of 256, 512, and 1,024 channels with residual concat--MLP fusion on Qwen3.5-0.8B.}

{The fused $C=256$ model matches $C=1024$ at 44.28\% on ScreenSpot-Pro and trails it by only one correct prediction on ScreenSpot-v2 (89.86\% versus 89.94\%), while using 51.3\% fewer total trainable parameters (Table~\ref{tab:parameter_allocation}). The $C=512$ model reduces total trainable parameters by 40.0\%, but its ScreenSpot-v2 and ScreenSpot-Pro scores are below those of $C=256$. In these runs, accuracy therefore does not improve uniformly with decoder width. The $C=1024$ configuration gives the highest ScreenSpot and ScreenSpot-v2 accuracy, whereas $C=256$ offers a smaller parameter budget while retaining the same ScreenSpot-Pro accuracy.}

\FloatBarrier

\section{{Measured Inference Cost}}
\label{app:efficiency}
The two spatial decoder convolutions require $18hwC^2$ multiply--accumulates per query, and the output projection requires $hwCs^2$; pixel shuffle only rearranges values.
Thus spatial refinement operates on $h\times w$, before producing the $H'\times W'$ action field.

{Table~\ref{tab:inference_cost} reports inference cost for Qwen2.5-VL-3B on one NVIDIA H200 with batch size 1. Each run covers the full benchmark: 1,272 examples for ScreenSpot and ScreenSpot-v2, and 1,581 for ScreenSpot-Pro. Mask prediction and click extraction together average 42.49--63.59\,ms per example and account for 8.25--11.62\% of evaluation-loop wall time.}

\begin{table}[!htbp]
\centering
\fontsize{8}{9.6}\selectfont
\setlength{\tabcolsep}{2pt}
\caption{{Inference cost of Qwen2.5-VL-3B on one H200 with batch size 1. Decoder + Post. gives combined mask prediction and click extraction time, its share of wall time, and mean time per example. Peak memory is allocated GPU memory. Derived values use unrounded measurements.}}
\label{tab:inference_cost}
\begin{tabular*}{\linewidth}{@{\extracolsep{\fill}}lrrrrrrrrr@{}}
\toprule
Dataset & \multicolumn{5}{c}{{Stage time (s)}} & \multicolumn{3}{c}{{Decoder + Post.}} & {Peak memory}\\
\cmidrule(lr){2-6}\cmidrule(lr){7-9}
& Wall & Prep. & Backbone & Decoder & Post. & {Time (s)} & {Share (\%)} & {ms/example} & {(GiB)}\\
\midrule
ScreenSpot & 472.5 & 173.3 & 232.4 & 7.4 & 46.7 & 54.04 & 11.44 & 42.49 & 8.60\\
ScreenSpot-v2 & 471.5 & 173.1 & 231.3 & 7.3 & 47.5 & 54.80 & 11.62 & 43.08 & 8.48\\
ScreenSpot-Pro & 1218.9 & 490.5 & 487.5 & 13.8 & 86.7 & 100.53 & 8.25 & 63.59 & 8.61\\
\bottomrule
\end{tabular*}
\end{table}

{Wall time covers the evaluation loop, including image and result I/O, and excludes setup and final reporting. Decoder time includes feature fusion and mask prediction; postprocessing extracts connected components and the click centroid on the CPU. Unassigned I/O and loop work account for the difference between wall time and the sum of stages.}

\FloatBarrier

\section{\revision{Qualitative Analysis and Limitations}}
\label{app:qualitative}

\subsection{{Strengths in Small-Target Grounding}}
\label{app:small_icons}

\begin{figure}[!htbp]
\centering
\includegraphics[width=\linewidth]{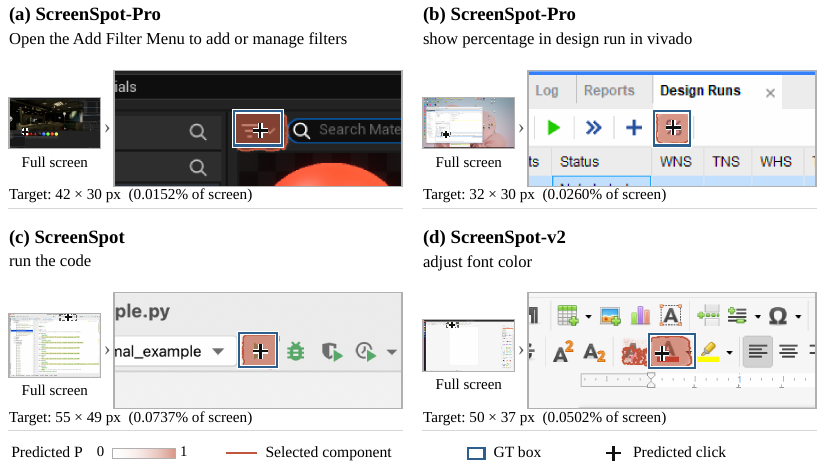}
\caption{Seg2Click-4B localizes small icons and adjacent controls through learned action masks. Warm overlays show predicted probabilities; outlines mark selected components, blue boxes mark targets, and crosses mark clicks. Insets are display-only enlargements.}
\label{fig:qualitative}
\end{figure}

{Figure~\ref{fig:qualitative} shows how Seg2Click-4B localizes compact controls in crowded interfaces. The Unreal Engine filter control and Vivado percentage icon measure only $42\times30$ and $32\times30$ pixels. In both cases, the predicted component concentrates on the target and its centroid yields a valid click. The code-execution and font-color examples similarly show localization among neighboring toolbar controls. These cases illustrate the fine spatial predictions enabled by action masks, using one prediction on the full screenshot without a verifier or an additional zoom-in pass.}

{The spatial output also makes localization behavior inspectable: the mask shows where the model responds, while the click shows how the deterministic rule converts that response into an action. This distinction helps diagnose whether a miss arises from a poorly localized region or from converting that region into a click. Enlarged views are for display only.}

\subsection{Failure Cases and Limitations}
\label{app:failure_cases}
\begin{figure}[H]
\centering\small
\includegraphics[width=0.92\linewidth]{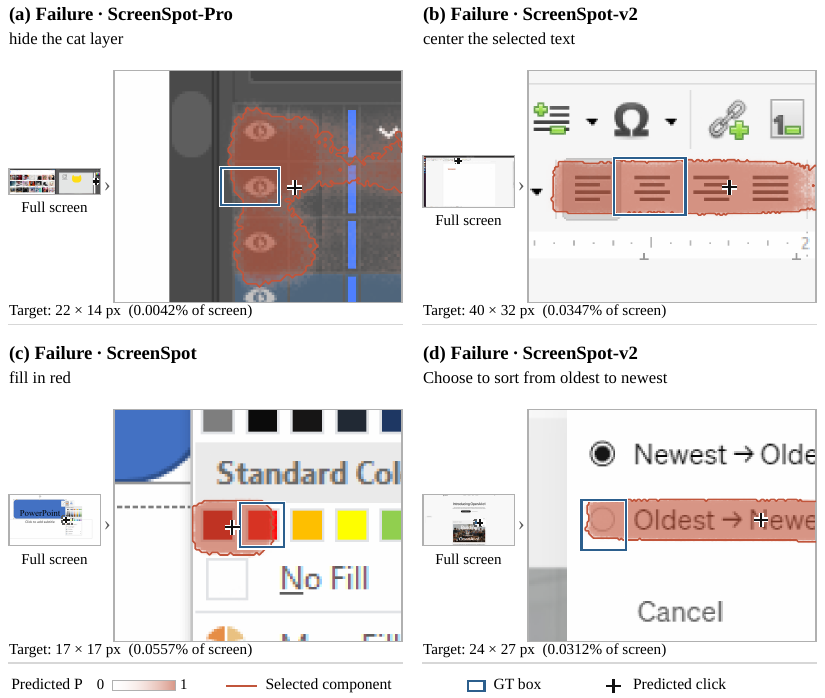}
\caption{Failures with responses extending beyond the annotated target: (a) layer visibility, (b) paragraph alignment, (c) color selection, and (d) a radio control and its label. Warm mask overlays and component outlines show how the centroid can fall outside the blue target box. Failure labels follow the same inclusive box-hit criterion as the quantitative evaluation.}
\label{fig:qualitative_extra1}
\end{figure}

{Figure~\ref{fig:qualitative_extra1} shows four failures in which the predicted region extends beyond the intended target, revealing limitations in boundary localization and mask-to-click decoding.}

\paragraph{{Separating adjacent controls.}}
{In Figure~\ref{fig:qualitative_extra1}(a--c), the selected component spans neighboring layer controls, alignment buttons, or color swatches. In (d), it covers both the radio button and its text label. The response reaches the relevant control group but does not isolate the annotated element. Fine output resolution therefore provides the capacity to separate nearby controls, but accurate boundaries still depend on the learned features and supervision.}

\paragraph{{Sensitivity of centroid selection.}}
{The largest-component rule selects one connected region, and its unweighted centroid gives every foreground pixel equal influence. When the region extends onto an adjacent control or label, these extra pixels can shift the click outside the target box even when the mask overlaps the correct element. This exposes a limitation of fixed geometric decoding: it does not assess whether the selected point lies in the intended control. Confidence-aware click selection could address this issue, but its benefit requires evaluation.}

\paragraph{{Scope of the evidence.}}
{These examples illustrate behavior rather than the frequency of each failure mode. Bounding-box supervision and evaluation approximate the valid click region; for instance, the label click in Figure~\ref{fig:qualitative_extra1}(d) fails the box-hit criterion, but a static screenshot cannot establish whether the interface would accept it. Evaluating Seg2Click within interactive agents remains necessary to measure task completion and recovery from grounding errors.}

\end{document}